\pdfoutput=1

\documentclass[11pt]{article}

\usepackage[final]{acl}

\usepackage{times}
\usepackage{latexsym}
\usepackage{xurl}
\usepackage{comment}
\usepackage{midfloat}

\usepackage[utf8]{inputenc}

\usepackage{microtype}

\usepackage{inconsolata}

\usepackage{graphicx}

\usepackage{xcolor}

\usepackage{colortbl}
\usepackage{booktabs}
\usepackage{multirow}
\usepackage[bottom]{footmisc}
\usepackage{amsmath}

\title{Breaking Language Barriers: Equitable Performance in Multilingual Language Models\textsuperscript{*}}

\author{
    Tanay Nagar$^{1,2}$\textsuperscript{\footnotemark[3]} \quad
    Grigorii Khvatskii$^{3}$ \quad
    Anna Sokol$^{3}$ \quad
    Nitesh V. Chawla$^{2,3}$ \\
    \\
    $^{1}$University of Wisconsin--Madison \\
    $^{2}$NSF Center for Computer Assisted Synthesis (CCAS), University of
  Notre Dame \\
    $^{3}$University of Notre
  Dame \\
    \texttt{tpnagar@wisc.edu} \quad \texttt{\{gkhvatsk, asokol,
  nchawla\}@nd.edu}
  }

\begin{document}
\maketitle

\begingroup
  \renewcommand\thefootnote{\fnsymbol{footnote}}
  \footnotetext[1]{Accepted as a non-archival work-in-progress paper at the Student Research Workshop (SRW), NAACL 2025.}
  \footnotetext[3]{Work was done when TN was a DATA SURF Fellow at the NSF Center for Computer Assisted Synthesis (CCAS), University of Notre Dame.}
\endgroup

\begin{abstract}

Cutting-edge LLMs have emerged as powerful tools for multilingual communication and understanding. However, LLMs perform worse in Common Sense Reasoning (CSR) tasks when prompted in low-resource languages (LRLs) like Hindi or Swahili compared to high-resource languages (HRLs) like English. Equalizing this inconsistent access to quality LLM outputs is crucial to ensure fairness for speakers of LRLs and across diverse linguistic communities. In this paper, we propose an approach to bridge this gap in LLM performance. Our approach involves fine-tuning an LLM on synthetic code-switched text generated using controlled language-mixing methods. We empirically demonstrate that fine-tuning LLMs on synthetic code-switched datasets leads to substantial improvements in LRL model performance while preserving or enhancing performance in HRLs. Additionally, we present a new dataset of synthetic code-switched text derived from the CommonSenseQA dataset, featuring three distinct language ratio configurations.\footnote{The code and data for this paper can be accessed through this github repo: \url{https://github.com/tnagar72/Breaking-Language-Barriers-Equitable-Performance-in-Multilingual-Language-Models}}

\end{abstract}

\section{Introduction}



The remarkable capabilities of LLMs for a wide range of language processing tasks have led to their use across countless fields and domains globally. However, the performance of LLMs is heavily influenced by the availability of textual data in different languages, impacting their overall effectiveness. For example,~\citet{li2024quantifying} demonstrated that existing LLMs show a noticeable performance gap between HRLs and LRLs. In CSR tasks across different languages, LLMs have been shown to have a performance gap of over 15\% on average \cite{zhang2023don}. This performance disparity arises due to an imbalance in training data availability for different languages. This can exacerbate the digital divide by limiting access to LLMs for LRLs, disproportionately affecting underrepresented communities.


Studies show that existing multilingual LLMs often either rely on a single dominant language or have separate internal representations of different languages \cite{zhong2024englishcentricllmslanguagemultilingual}. This can lead to the presence of deeply rooted linguistic biases in the model output \cite{demidova-etal-2024-john}. Considering that CSR tasks are based on the shared implicit human knowledge about everyday situations, biases can skew the model’s interpretation of diverse cultural contexts \cite{li-etal-2022-systematic}. In this project, we draw attention to the fact that in bilingual humans, lexicons of different languages have similar representations \cite{fabbro2001bilingual}. In recent literature, many techniques for cross-language adaptation of LLMs have been proposed \cite{yamaguchi2024empiricalstudycrosslingualvocabulary, fujii2024continualpretrainingcrosslingualllm,yamaguchi2024effectivelyexpandvocabularyllms,lin2024mala500massivelanguageadaptation}. However, to our knowledge, none of them have been designed to address this language representation challenge.

Prior research \cite{guo2023analyzingreducingperformancegap} has also shown that fine-tuning multilingual models exclusively on LRL data typically results in significant performance degradation in high-resource languages. This occurs due to the finite capacity of language models to represent multiple languages simultaneously, often leading to an undesirable trade-off where improving performance in LRLs would come at the cost of degraded HRL performance. Code-switching, the practice of alternating between multiple languages, offers a promising avenue for tackling this key problem. Code-switching could allow for a more equitable representation of both HRLs and LRLs, helping us move toward compound multilingual understanding in LLMs, which would bring forth a unified representation of knowledge across languages.

To summarize, our paper makes the following key contributions:

\begin{itemize}
    \item We demonstrate a performance gap in CSR tasks between Hindi and English in existing LLMs.
    \item  We develop and release a Hindi-English synthetic code-switched dataset
    \item We demonstrate that fine-tuning an existing LLM on a code-switched dataset results in a significant improvement in LRL performance without degrading performance on HRLs.
\end{itemize}

\section{Background}

\begin{figure*}[ht]
  \centering
\includegraphics[width=0.75\textwidth]{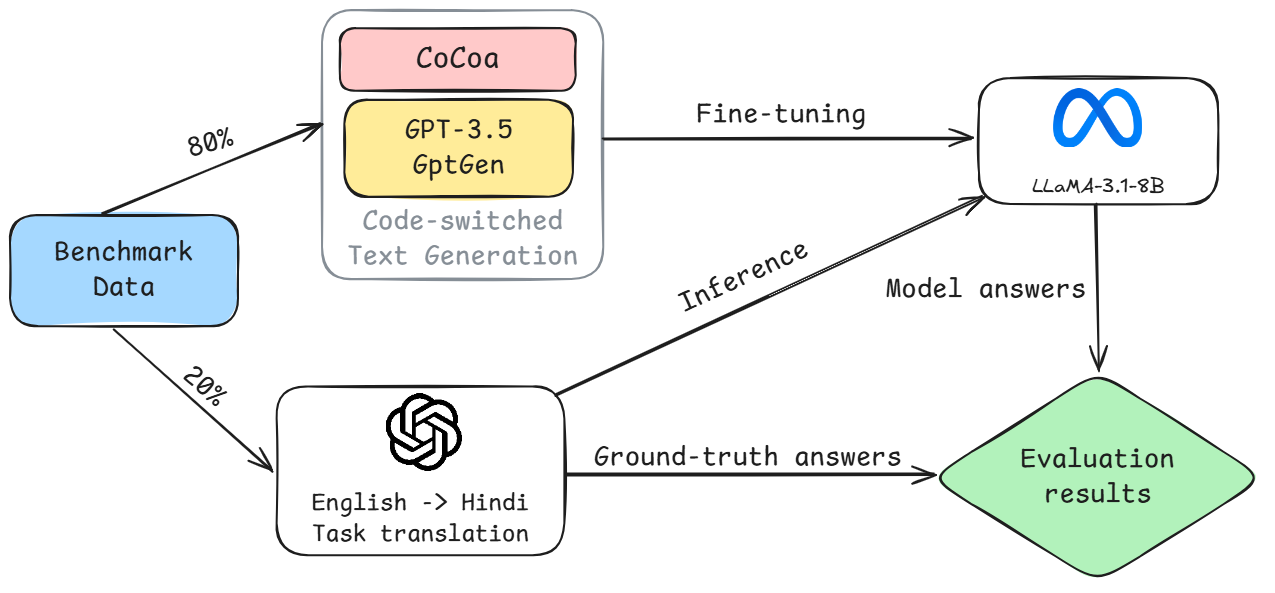}
  \caption{Overview of the experimental pipeline}
  \label{fig:experiments}
\end{figure*}

The study of bilingualism and multilingualism has long been a topic of interest for researchers, as it offers insights into the mechanisms underlying language processing and acquisition \cite{https://doi.org/10.1111/lang.12529, fricke2019bilingualism}. The advent of LLMs has not only increased this interest but also presented new challenges for these tasks, revealing a growing disparity in how language technologies handle diverse linguistic needs. This discrepancy has implications for language accessibility, and the ability of underrepresented communities to benefit from AI advancements.

The use of linguistically diverse prompts has already been shown to improve LLM performance across a variety of tasks \cite{nguyen2024democratizingllmslowresourcelanguages}. Leveraging code-switching is a gradual next step to enhance LLM representation and performance on LRLs. Code-switching is a natural phenomenon that occurs in multilingual communities, where speakers alternate between two or more languages in the same sentence during communication. This practice usually involves alternating between a matrix language L1 and a dominant language L2. 

The practice of code-switching can enrich language models by exposing them to mixed linguistic structures and semantics, thereby improving the model's robustness and adaptability in multilingual contexts. However, naturally occurring code-switched datasets are scarce, particularly for LRLs, which presents a significant challenge for training models effectively on such data \cite{9074205, yong2023promptingmultilinguallargelanguage}, thus underscoring the need for generating synthetic code-switched text instead.

Recent advances in controlled text generation techniques have opened new opportunities for synthesizing high-quality code-switched data. For example, CoCoa \cite{mondal-etal-2022-cocoa} allows calibration over semantic properties, such as the frequency of switching between languages, as well as setting the ratio between them in the resulting text. This level of control can help evaluate how different properties of the synthetic code-switched text affect downstream LLM performance. This control is crucial for creating synthetic datasets that can be used to systematically explore the effects of varying levels of code-switching on LLM training and performance.

Additionally, open-source LLMs have demonstrated potential in generating coherent and contextually rich text \cite{yong2023promptingmultilinguallargelanguage}, making them a useful tool for augmenting the training datasets for LRLs through synthetic code-switching. 



In this paper, we evaluate three variants of this dataset, employing three distinct ratios between languages. Finally, we show empirically that fine-tuning an existing LLM on a synthetic code-switched dataset leads to improved performance for LRLs with little to no degradation for HRLs. Our work can thus serve as a foundation for building future LLMs that offer state-of-the-art performance in LRLs, as well as more equitable language representation.

\section{Methods}

In this section, we present our methodology to mitigate the performance gap between HRLs and LRLs through two main steps: (1) generating synthetic code-switched datasets and (2) fine-tuning LLMs with this augmented data. The overview of our pipeline can be found in Figure \ref{fig:experiments}.

\subsection{Synthetic Code-Switched Text Generation}


Using synthetic code-switched text generation methods, we aimed to produce coherent, well-structured sentences that accurately reflect natural code-switching in multilingual communities. For this purpose, we employed two approaches for data generation: using a large pre-trained LLM and, the CoCoa model \cite{mondal-etal-2022-cocoa}.

We used GPT-3.5 \cite{brown2020languagemodelsfewshotlearners} to generate code-switched text by creating a detailed prompt, instructing the conversion of English statements into Hinglish (a mix of Hindi and English). Specifically, the prompt instructed the model to write Hindi words in Devanagari script and English words in Latin script, aiming to create a balanced and natural blend of both languages in each sentence. We also included some few-shot examples to illustrate the desired style of code-switching, hoping to guide the model toward more naturally coherent outputs.

Despite multiple efforts, GPT-3.5 could not effectively control language-mixing ratios. Requests for specific language ratios resulted in inconsistent outputs, often skewed heavily towards English or generating several distinct English and Hindi sentences, with minimal code-switching in most sentences. We called the dataset generated using this process GPTgen.

To achieve precise control over language mixing ratios, we utilized a simpler variant of the CoCoa model (300M parameters, model weights provided by the authors), which allowed for adjusting the Code-Mixing Index (CMI), a measure of mixing between the languages used.

The CMI is calculated based on the proportion of words from each language (L1 and L2) used in a given text, weighted by their frequency and distribution across sentences. Formally, \citet{das-gamback-2014-identifying} define it as:
\begin{equation}
    \text{CMI} = 
    \begin{cases} 
        100\% \times \left[ 1 - \frac{\max\{w_i\}}{n - u} \right] & : n > u \\
        0 & : n = u 
    \end{cases}
\end{equation}
where  $w_i$ is the number of words from a particular language, $\max\{w_i\}$ is the highest number of words in any language, $n$ is the total number of tokens, and $u$ is the number of language-independent tokens. This formula results in a value between 0\% (no code-switching, monolingual text) and 50\% (maximum code-switching, an equal mix of all languages involved). 

\renewcommand{\arraystretch}{1.3}  
\begin{table*}[h]
\centering
\resizebox{\textwidth}{!}{%
\begin{tabular}{|l|c|c|c|c|c|c|c|c|c|c|}
\hline
\textbf{} & \multicolumn{2}{c|}{\textbf{Baseline}} & \multicolumn{2}{c|}{\cellcolor[HTML]{E7EFDD}\textbf{GPTgen}} & \multicolumn{2}{c|}{\cellcolor[HTML]{CCDAF6}\textbf{CMI 1}} & \multicolumn{2}{c|}{\cellcolor[HTML]{D8D2E7}\textbf{CMI 2}} & \multicolumn{2}{c|}{\cellcolor[HTML]{F8E6D0}\textbf{CMI 3}} \\ \hline
          & \textbf{English} & \textbf{Hindi}  & \textbf{English} & \textbf{Hindi} & \textbf{English} & \textbf{Hindi} & \textbf{English} & \textbf{Hindi} & \textbf{English} & \textbf{Hindi}  \\ \hline
\textbf{Mean Accuracy} & 78.00\%& 54.00\%& 88.80\%& 79.60\%& 81.60\%& 75.20\%& \textbf{90.40\%}& \textbf{85.60\%}& 87.20\%& 77.20\% \\ \hline
\textbf{Std Dev (\%)} & & & 6.26\% & 12.76\% & 14.72\% & 16.16\% & 2.97\% & 3.29\% & 4.15\% & 8.32\% \\ \hline
\end{tabular}%
}
\caption{Average Accuracy results across models along with baselines scores. The highest values are in bold.}
\label{table1}
\end{table*}
\renewcommand{\arraystretch}{1.0}  

In our work, we generated text with three specific CMI ranges: low (from 0 to  16.7\%), medium (16.7\% to 30\%), and high (30\% to 50\%). These three ranges were created to aid in a better understanding of how language ratios affect the final result. The 50\% maximum is set, since above this threshold, the dominant and matrix languages get switched, replicating scenarios that were already considered in CMI $\le 50\%$.

This fine-grained control was essential for creating datasets that reflect varying degrees of code-switching intensity, aiding in a better understanding of how different ratios affect the final result. The datasets generated using this approach were the CMI 1 (low), CMI 2 (medium), and CMI 3 (high) datasets, corresponding to the three language mixing ratios mentioned above.

\subsection{Dataset Preparation}

We transformed the original English questions into code-switched Hindi-English text using the methods outlined in the Data Generation section. We ensured that the answer choices remained in English, focusing the code-switching transformation only on the questions. By maintaining the answers in English, we leverage the model’s strong foundational understanding of English semantics, aiming to transfer this understanding to the target language (Hindi) through fine-tuning. This process led to the creation of four distinct datasets. The commonSenseQA is a widely accepted dataset, and we rely on the evaluation metrics released with the CoCoa paper to support the reliability of the generated dataset. Additionally, we conducted a manual verification process by reviewing one randomly selected question from each batch of 50 questions in the 1,200-question dataset to ensure multilingual coherence. Examples of original questions and their code-switched versions generated using each of the four settings can be found in Appendix \ref{ap:examples}.

\subsection{Fine-Tuning Process}

We utilized the \texttt{LLaMA-3-8B-Instruct} (8B parameters, available under the LLaMA 3 CLA) model developed by Meta AI \cite{dubey2024llama3herdmodels} as the base model for our fine-tuning experiments. We selected this model due to its availability for research and proven effectiveness in multilingual contexts. Its tokenizer supports both Devanagari and Latin scripts used in Hindi and English, respectively. This feature minimized the need for complex preprocessing steps to handle code-switched inputs.

\section{Experiments}

In this section, we elaborate on the tests conducted to evaluate our fine-tuned \texttt{LLaMA-3-8B-Instruct} model on CSR tasks. 

\subsection{Dataset}

We used our aforementioned data generation methods to augment an existing English-language dataset called CommonSenseQA~\cite{talmor2018commonsenseqa} (available under the MIT license) and create a new dataset. CommonSenseQA contains 12,102 multiple-choice questions designed to test commonsense reasoning. We focus on this dataset as it provides us with an opportunity to test the model performance on questions that require a significant degree of language understanding but where the answer does not depend on the language of the question. This makes this dataset an ideal candidate for evaluating LLM CSR capabilities. This dataset is widely used for evaluating language models' CSR performance\cite{zhao2024surveylargelanguagemodels, srivastava2023imitationgamequantifyingextrapolating,zhang2023don}.

\subsection{Experimental setup and evaluation metrics}


To reduce the effects of data partitioning, we employ a five-fold cross-validation method for testing (see Appendix \ref{ap:expr} for per-fold results). To assess the performance of our fine-tuned LLM on the testing dataset, we used accuracy, calculated as the proportion of correctly answered questions out of the total number of questions in the test set. Since our inference procedure was non-deterministic, we presented each multiple-choice question to our fine-tuned model five times and used the most frequent output for evaluation. The model was instructed to respond in a specified way to all questions and to pick the right option apart from the four distractor options. 

We also limited the output length to focus the model on producing a single, coherent answer per question to prevent multiple answers and maintain clarity in the evaluation. The same testing dataset was also translated into Hindi to assess the performance gap for our LLM, and the language accuracy for the Hindi and English versions of each question was calculated. Along with evaluating the models fine-tuned on our four distinct datasets, we also similarly calculate baseline scores with the base model to understand performance changes because of our fine-tuning step.

All training and inference was conducted on compute nodes with 256GB RAM, Intel Xeon Platinum 8358 CPU, and 8 NVIDIA A100 (80GB VRAM) or 8 NVIDIA H100 (80GB VRAM) GPUs. We conducted our experiments using the PyTorch framework for model inference and fine-tuning. The models were fine-tuned over 5 epochs using a learning rate of $3\cdot10^{-5}$ and a batch size of 32, employing the Adam optimizer and utilizing QLoRA \cite{dettmers2023qloraefficientfinetuningquantized} to reduce memory overhead.

\section{Preliminary Results}

In this section, we present the empirical findings of our experiments, elucidating the impact of fine-tuning LLMs on synthetic code-switched datasets with varying CMIs. Table~\ref{table1} summarizes the mean accuracies achieved by the models across different configurations. 

Our results indicate that fine-tuning the LLM on synthetic code-switched datasets significantly enhances its performance on Hindi tasks while maintaining or even improving accuracy on English tasks. Notably, the model fine-tuned with the \textbf{CMI 2} dataset shows better performance, achieving an average accuracy of \textbf{90.40\%} on English and \textbf{85.60\%} on Hindi tasks. 

The superiority of the CMI 2 configuration can be attributed to its optimal balance in code-mixing intensity. The medium CMI 2 introduces a harmonious blend of linguistic elements from both English and Hindi, facilitating more effective cross-linguistic transfer and representation learning within the model. It is curious that this mirrors a result known from human experimentation, where moderate levels of bilingualism were shown to improve human performance in their native language \cite{Grosjean_2015, dijkstra_architecture_2002}.

From a linguistic standpoint, moderate code-switching mirrors natural bilingual discourse, where speakers fluidly alternate between languages without reliance on either. This balanced code-mixing enables the model to capture nuanced syntactic structures and semantic relationships that are characteristic of both languages, thereby enriching its overall language understanding capabilities.

\section{Limitations}
Our method is currently evaluated on a single language pair, Hindi-English.  Future research should expand on these experiments to include additional low-resource languages and diverse linguistic families to validate the generalizability of our findings.

Our study was limited by the models we used for synthetic code-switched text generation. In the future, we plan to include more modern generation techniques like GPT-4o into our pipeline. Our experimental results were also limited by the relatively smaller cross-validation folds we analyzed.  

Another limitation relates to the models we used for data synthesis. For example, the authors of the CoCoa model state that the model may have difficulty scaling to long sentences. These limitations can, in turn, propagate to our fine-tuned models.  Additionally, the CoCoa model outputs may still not completely encompass the natural nuances of spontaneous human code-switching. A particular risk is that biases present in code-switched text generation models can propagate into our fine-tuned models as well. Although we employed controlled language mixing, there are limitations on how well synthetically generated data models real-world scenarios.

Finally, our evaluation metrics focused primarily on accuracy in CSR and CMI. A more comprehensive evaluation involving diverse metrics, including additional tasks,  will be more useful in getting a better understanding of the effects of such fine-tuning on overall model performance.

\section{Discussion}

Despite inconsistencies in language mixing during data generation, the GPTgen dataset still lead to noticeable performance gains. This suggests that any degree of code-switching can enhance multilingual performance and encourage learning of cross-linguistic representations, even if code-switching patterns aren't strictly controlled.

Our experiments maintained the answer choices in English, while code-switching only the questions. However, utilizing fully code-switched datasets (both answers and questions) could provide additional insights into the model's robustness and real-world alignment. Exploring this will help understand whether full code-switched datasets lead to improved cross-lingual transfer or lead to semantic misalignment.

\section{Conclusion and Future directions}

This work shows code-switched fine-tuning as a promising approach to improving LRL performance while preserving/enhancing HRL performance. Our results suggest that this approach is a much more balanced alternative to monolingual fine-tuning, thus mitigating the issues of catastrophic forgetting that occurs when LLMs are trained exclusively on LRL data. 
Our current work is in progress. Future work will explore how these findings generalize to other languages, especially Russian- and Spanish-English language pairs. Further, we plan to extend this methodology to two additional LLMs—Qwen 2.5-7B \cite{qwen2025qwen25technicalreport} and Phi-3.5-mini \cite{abdin2024phi3technicalreporthighly}—and two additional benchmarks: XCOPA \cite{ponti-etal-2020-xcopa} and OpenBookQA \cite{mihaylov2018suitarmorconductelectricity}. While naturally-occurring code-switched datasets are scarce, particularly for LRLs, our anticipated work will also explore augmenting our training data by incorporating real code-switched datasets, such as those presented in the LinCE benchmark \cite{aguilar-etal-2020-lince}.

We also plan to benchmark our approach against models fine-tuned on fully translated monolingual datasets to contrast the specific effects of code-switching from direct target-language exposure. Finally, we intend to experiment with more precise control over code-mixing indexes and fully code-switched datasets to understand how these could further optimize multilingual model adaptation.


\bibliography{bibliography/custom}

\appendix

\renewcommand{\thetable}{\Alph{section}\arabic{table}}

\clearpage

\section{Example synthetic code-switched questions}
\label{ap:examples}
\setcounter{table}{0}

We provide three examples of questions from the commonSenseQA dataset, as well as their corresponding code-switched versions. The examples are provided in Table \ref{tab:synthexamples}.

\begin{strip}
    \centering
    \captionsetup{type=table}
    \includegraphics[width=0.9\textwidth]{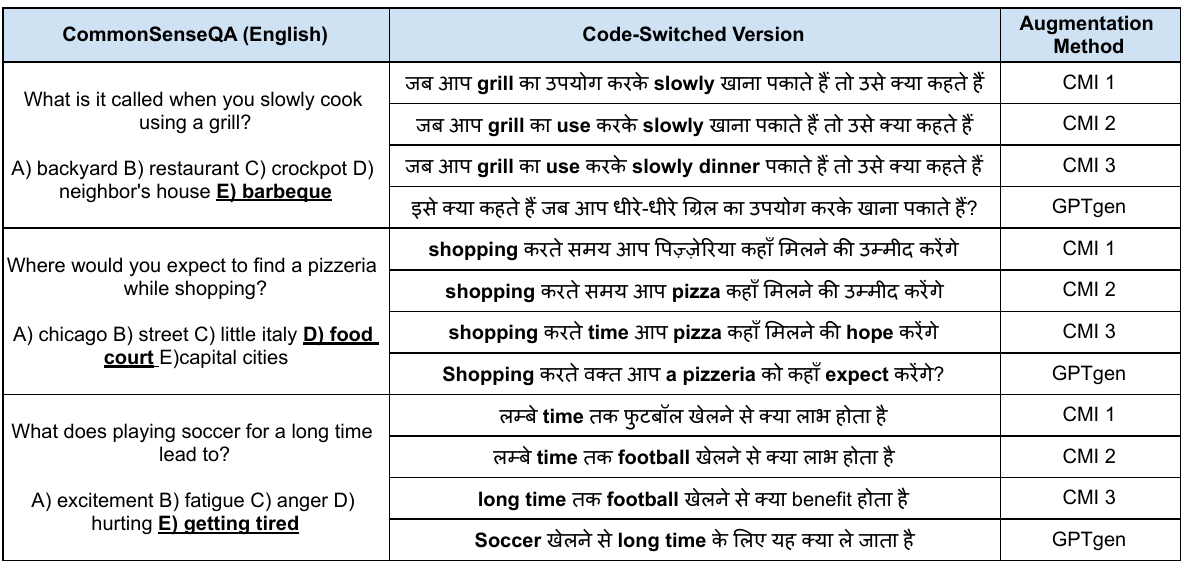}
    \caption{Examples of synthetic code-switched questions. Correct answers are \underline{\textbf{bold-underlined}}}
    \label{tab:synthexamples}
\end{strip}


\section{Per-fold table of experimental results}
\label{ap:expr}

We provide a table of evaluation results for each of the five cross-validation folds. The results are provided in Table \ref{tab:foldperf}.

\begin{strip}
    \centering
    \captionsetup{type=table}
    \includegraphics[width=1\textwidth]{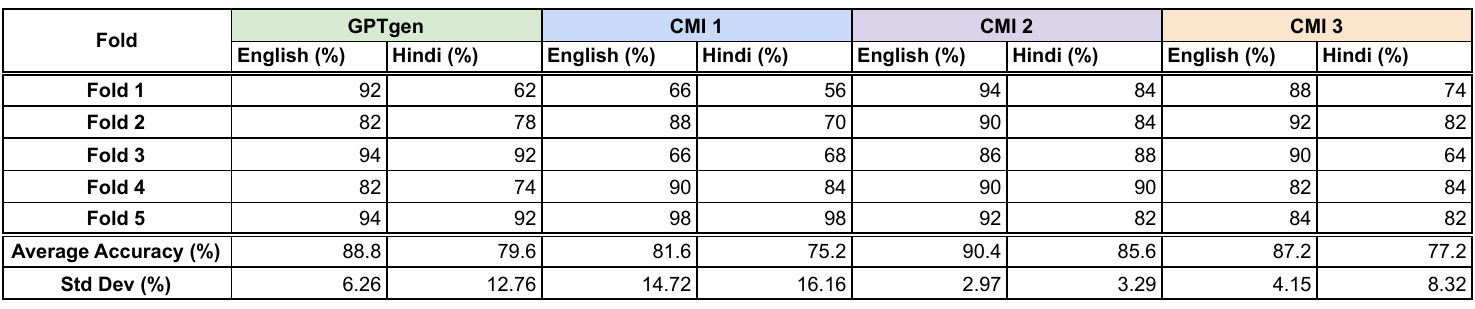}
    \caption{Performance comparison across folds and language configurations, including standard deviations in percentage.}
    \label{tab:foldperf}
\end{strip}
\end{document}